\documentclass{article}
\usepackage{ijcai20}

\usepackage{times}
\usepackage{soul}
\usepackage{url}
\usepackage[utf8]{inputenc}
\usepackage[small]{caption}
\usepackage{graphicx}
\usepackage{amsmath}
\usepackage{amsthm}
\usepackage{booktabs}
\usepackage{algorithm}
\usepackage{algorithmic}
\usepackage{multirow}
\usepackage{array}

\usepackage{amsfonts}

\newcommand{\tabincell}[2]{\begin{tabular}{@{}#1@{}}#2\end{tabular}}

\title{SBAT: Video Captioning with Sparse Boundary-Aware Transformer}

\author{
Tao Jin$^1$
,
Siyu Huang$^2$,
Ming Chen$^{3}$,
Yingming Li$^1$\thanks{Corresponding author.},
Zhongfei Zhang$^4$
\affiliations
$^1$College of Information Science \& Electronic Engineering, Zhejiang University, China\\
$^2$Baidu Research, China\\
$^3$Alibaba Group, China\\
$^4$Department of Computer Science, Binghamton University, USA
\emails
\{jint\_zju,yingming\}@zju.edu.cn,
huangsiyu@baidu.com,
black.cm@alibaba-inc.com,
zzhang@binghamton.edu
}

\begin{document}

\maketitle

\begin{abstract}
 In this paper, we focus on the problem of applying the transformer structure to video captioning effectively. The vanilla transformer is proposed for uni-modal language generation task such as machine translation. However, video captioning is a multimodal learning problem, and the video features have much redundancy between different time steps. Based on these concerns, we propose a novel method called sparse boundary-aware transformer (SBAT) to reduce the redundancy in video representation. SBAT employs boundary-aware pooling operation for scores from multihead attention and selects diverse features from different scenarios. Also, SBAT includes a local correlation scheme to compensate for the local information loss brought by sparse operation. Based on SBAT, we further propose an aligned cross-modal encoding scheme to boost the multimodal interaction. Experimental results on two benchmark datasets show that SBAT outperforms the state-of-the-art methods under most of the metrics.
\end{abstract}

\section{Introduction}

Recently, the combination of vision and language attracts more and more attention \cite{you2016image,pan2017video,antol2015vqa,li2019beyond}. Video captioning is a valuable but challenging task in this topic, where the goal is to generate text descriptions for video data directly. The difficulties of video captioning mainly lie in the modeling of temporal dynamics and the fusion of multiple modalities. 

Encoder-decoder structures are widely used in video captioning \cite{shen2017weakly,aafaq2019spatio,pei2019memory,wang2019controllable,gan2017semantic}. In general, the encoder learns multiple types of features from raw video data. The decoder utilizes these features to generate words. Most encoder-decoder structures are built upon the long short-term memory (LSTM) unit, however, LSTM has two main drawbacks. First, LSTM-based decoder does not allow a parallel prediction of words at different time steps, since its hidden state is computed based on the previous one. Second, LSTM-based encoder has insufficient capacity to capture the long-range temporal correlations.



\begin{figure}[t]
	\centering
	\includegraphics[scale=0.75]{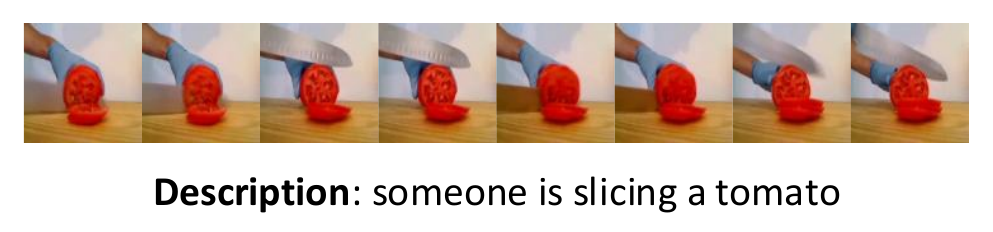}
	\vspace{-0.2cm}
	\caption{An example of redundancy between video frames.}
	\label{fig:redun} 
	\vspace{-0.5cm}
\end{figure}

To tackle these issues, \cite{chen2018tvt} and \cite{zhou2018end} proposed to replace LSTM with transformer for video understanding. Specifically, \cite{chen2018tvt} used multiple transformer-based encoders to encode video features and a transformer-based decoder to generate descriptions. Similarly, \cite{zhou2018end} utilized transformer for dense video captioning, \cite{zhou2018end} utilized a transformer-based encoder to detect action proposals and described them simultaneously with a transformer-based decoder. Different from LSTM, the self-attention mechanism in transformer correlates the features at any two time steps, enabling the global association of features. However, the vanilla transformer is limited in processing video features with much temporal redundancy like the example in Fig. \ref{fig:redun}. In addition, the cross-modal interaction between different modalities is ignored in the existing transformer-based methods.





Motivated by the above observations, we propose a novel method named sparse boundary-aware transformer (SBAT) to improve the transformer-based encoder and decoder architectures for video understanding. In the encoder, we employ sparse attention mechanism to better capture the global and local dynamic information by solving the redundancy between consecutive video frames. Specifically, to capture the global temporal dynamics, we divide all the time steps into $n$ chunks according to the gradient values of attention logits and select $n$ time steps with top-$n$ gradient values. To capture the local temporal dynamics, we implement self-attention between $r$ adjacent time steps. In the decoder, we also employ the boundary-aware strategy for encoder-decoder multihead attention. In addition, we implement cross-modal sparse attention following the self-attention layer to align multimodal features along temporal dimension. We conduct extensive empirical studies on two benchmark video captioning datasets. The quantitative, qualitative and ablation experimental results comprehensively reveal the effectiveness of our proposed methods. 



The main contributions of this paper are three-folded:

(1) We propose the sparse boundary-aware transformer (SBAT) to improve the vanilla transformer. We use boundary-aware pooling operation following the preliminary scores of multihead attention and select the features of different scenarios to reduce the redundancy.

(2) We develop a local correlation scheme to compensate for the local information loss brought by sparse operation. The scheme can be implemented synchronously with the boundary-aware strategy.


(3) We further propose a cross-modal encoding scheme to align the multimodal features along the temporal dimension.



\section{Related Work}


As a popular variant of RNN, LSTM is widely used in existing video captioning methods. \cite{venugopalan2015sequence} utilized LSTM to encode video features and decode words. \cite{yao2015describing} integrated the attention mechanism into video captioning, where the encoded features are given different attention weights according to the queries of decoder. \cite{hori2017attention} further proposed a two-level attention mechanism for video captioning. The first level focuses on different time steps, and the second level focuses on different modalities. \cite{long2018video} and \cite{jin2019recurrent} detected local attributes and used them as supplementary information. \cite{jin2019low} introduced cross-modal correlation into attention mechanism. Recently, \cite{chen2018tvt} proposed to replace LSTM with transformer in video captioning models. However, directly using transformer for video captioning has several drawbacks, i.e., the redundancy of video features and the lack of multimodal interaction modeling. In this paper, we propose a novel approach called sparse boundary-aware transformer (SBAT) to address these problems.    


\begin{figure*}[h]
	\centering
	\includegraphics[scale=0.75]{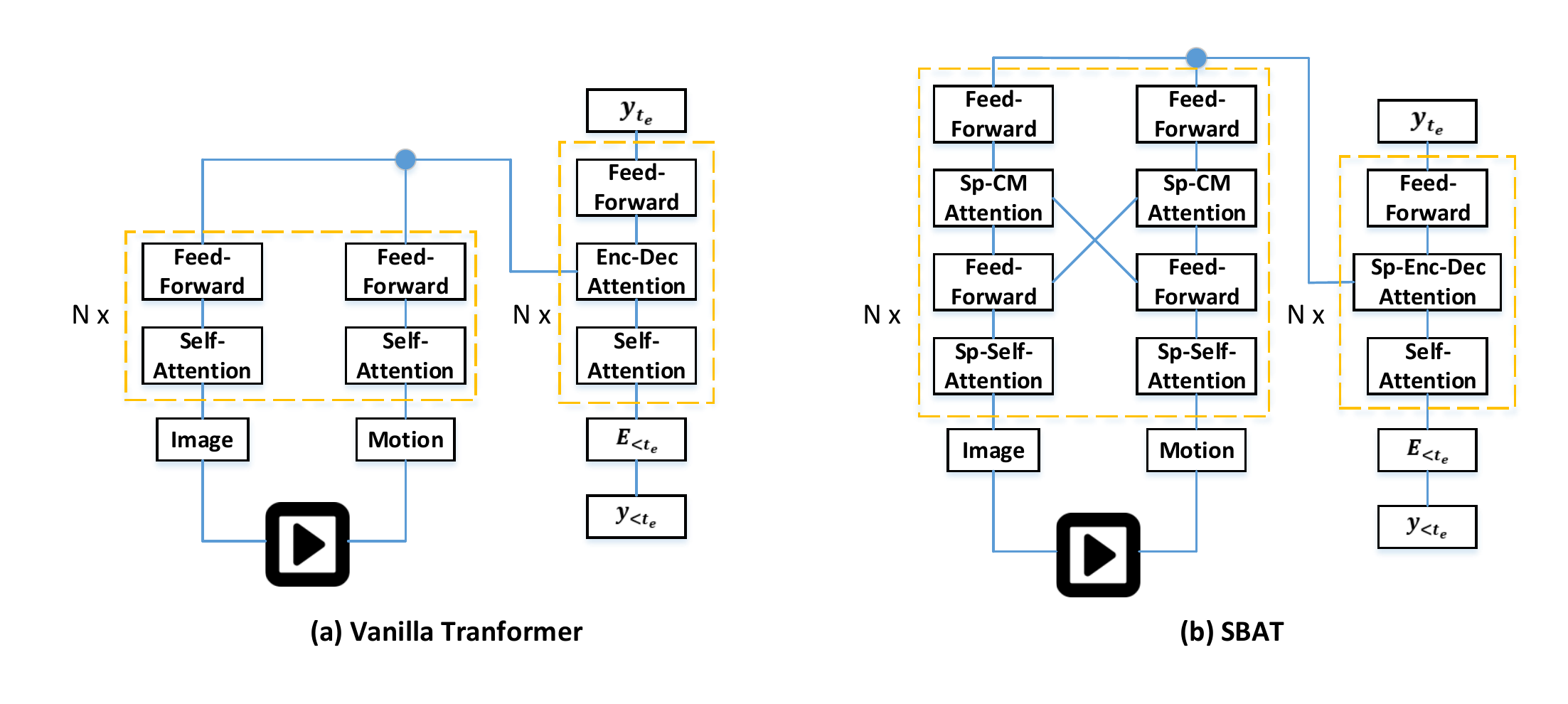}
	\vspace{-0.3cm}
	\caption{(a) is the overall framework of Vanilla Transformer. It consists of multihead attention mechanism and feed-forward neural network. The features of different modalities are processed separately and the queries of decoder associate these features to generate words. $N$ denotes the number of stacked blocks. (b) is the architecture of SBAT. It introduces sparse boundary-aware strategy (Sp) into all the multihead attention blocks in encoder and decoder. In addition, we learn cross-modal interaction after the first feed-forward layer in an encoder block.}
	\label{fig:overall} 
	\vspace{-0.4cm}
\end{figure*}

\section{Transformer-based Video Captioning}

Transformer \cite{vaswani2017attention} is originally proposed for machine translation. Due to the effectiveness and scalability, transformer is employed in many other tasks including video captioning. A simple illustration of transformer-based video captioning model is shown in Fig. \ref{fig:overall}(a). The encoder and decoder both consist of multihead attention blocks and feed-forward neural network.

\subsection{Encoder} 

Different from the uni-modal inputs of machine translation, the inputs of video captioning are typically multimodal. As shown in Fig. \ref{fig:overall}(a), two separate encoders process image and motion features, respectively. We use $I \in \mathbb{R}^{T_i \times d}$ and $M \in \mathbb{R}^{T_m \times d}$ to denote the image and motion features, respectively. Here we take the process of image encoding as an example. The self-attention layer is formulated as
\begin{equation}
\label{1}
{\rm SelfAttention}(I) = {\rm MultiHead}(I,I,I)
\end{equation}
\vspace{-0.5cm}
\begin{equation}
\label{2}
{\rm MultiHead}(I,I,I) = {\rm Cat}({\rm head_1}, ..., {\rm head_h})W_1
\end{equation}

\noindent where "Cat" denotes concatenation operation, $W_1 \in \mathbb{R}^{d \times d}$ is a trainable variable. Multihead attention is a special variant of attention, where each head is calculated as
\begin{equation}
\label{3}
{\rm head_i} = {\rm Attention}(IW_i^Q, IW_i^K, IW_i^V)
\end{equation}

\noindent where $W_i^Q$, $W_i^K$, and $W_i^V \in \mathbb{R} ^ {d \times \frac{d}{h}}$ are also trainable variables, ``Attention" denotes scaled dot-product attention:
\begin{equation}
\label{4}
{\rm Attention}(Q, K, V) = {\rm softmax}(\frac{QK^ \mathbf{T}}{\sqrt{d}}) V
\end{equation}

\noindent where $d$ is dimension of $Q$ and $K$.
We adopt residual connection and layer normalization after the self-attention layer:

\begin{equation}
\label{5}
x = {\rm LayerNorm}(I + {\rm SelfAttention}(I))
\end{equation}
Every self-attention layer is followed by a feed-forward layer (FFN) that employs non-linear transformation: 
\begin{equation}
\label{6}
{\rm FFN}(x) = {\rm max}(0, xW_2 + b_2) W_3 + b_3
\end{equation}

\begin{equation}
\label{7}
I^{'} = {\rm LayerNorm}(x + {\rm FFN}(x))
\end{equation}


\noindent where $W_2 \in \mathbb{R}^{d \times 4d}$, $b_2 \in \mathbb{R}^{4d}$, $W_3 \in \mathbb{R}^{4d \times d}$, and $b_3 \in \mathbb{R}^{d}$ are trainable variables. The encoded image features $I^{'}$ is the output of an encoder block. The encoded motion features $M^{'}$ are calculated in the same way.


\subsection{Decoder}

The decoder block consists of self-attention layer, enc-dec attention layer, and feed-forward layer. In the self-attention layer, the word embeddings of different time steps associate with each other, and we take the output features as queries. In the enc-dec multihead attention layer, the query first associates image and motion features to get two context vectors respectively, then generates the words. The feed-forward layer in decoder is the same as Eqns. \ref{6} and \ref{7}. We also adopt residual connection and layer normalization after all the layers of the decoder.


Specifically, we use $E \in \mathbb{R}^{T_e \times d}$ to denote the embeddings of target words. To predict the word $y_{t_e}$ at time step $t_e$, the self-attention layer is formulated as
\begin{equation}
\label{8}
E_{<t_e}^{'} = {\rm LayerNorm}(E_{<t_e} + {\rm SelfAttention}(E_{<t_e}))
\end{equation}

\noindent where $E_{<t_e} \in \mathbb{R}^{(t_e -1) \times d}$ denotes the word embeddings of time steps less than $t_e$. The enc-dec attention layer is:
\begin{equation}
\label{9}
I_{t_e} = {\rm MultiHead}(E_{t_e {\text-} 1}^{'}, I^{'}, I^{'})
\end{equation}
\vspace{-0.3cm}
\begin{equation}
\label{10}
M_{t_e} = {\rm MultiHead}(E_{t_e {\text-} 1}^{'}, M^{'}, M^{'})
\end{equation}

\noindent Following \cite{hori2017attention}, we employ a hierarchical attention layer for $I_{t_e}$ and $M_{t_e}$:
\begin{equation}
\label{11}
V_{t_e} \!\! = \! {\rm LayerNorm}(E_{t_e{\text -}1}^{'} + {\rm MultiHead}(E_{t_e {\text -} 1}^{'},\! C_{t_e}, \! C_{t_e}\!))
\end{equation}
\vspace{-0.2cm}
\begin{equation}
\label{12}
C_{t_e} = [I_{t_e}, M_{t_e}]
\end{equation}

\noindent $V_{t_e}^{'} = {\rm LayerNorm}(V_{t_e} + {\rm FFN}(V_{t_e}))$ denotes the output of feed-forward layer. We calculate the probability distributions of words as:
\begin{equation}
\label{13}
Pr(y_{t_e}|y_{<t_e}, I^{'}, M^{'}) = {\rm softmax}(W_p V_{t_e}^{'} + b_p)
\end{equation}

\noindent where $I^{'}$ and $M^{'}$ denote the encoded video features. The optimization goal is to minimize the cross-entropy loss function defined as accumulative loss from all the time steps:

\begin{equation}
\label{14}
L=-\sum_{t_e=1}^{T_e} \log Pr(y_{t_e}^{*}|y_{<t_e}^{*}, I^{'}, M^{'})
\end{equation}

\noindent where $y_{t_e}^{*}$ denotes the ground-truth word at time step $t_e$.


\section{Sparse Boundary-Aware Attention}

Considering the redundancy of video features, it is not appropriate to compute attention weights using vanilla multihead attention. To solve the problem, we introduce a novel sparse boundary-aware strategy into the multihead attention. In Section \ref{4.1}, we introduce the sparse boundary-aware strategy. In Section \ref{4.2}, we provide the analysis of sparse boundary-aware strategy. In Section \ref{4.3}, we introduce the local correlation attention which compensates for the local information loss. In Section \ref{4.4}, we introduce an aligned cross-modal encoding scheme based on SBAT.

\subsection{Sparse Boundary-Aware Pooling}
\label{4.1}
We employ sparse boundary-aware strategy following the scaled dot-product attention logits. Specifically, the original logits are calculated as follows:



\begin{equation}
\label{15}
P = \frac{QK^\mathbf{T}}{\sqrt{d}}
\end{equation}

\noindent where $Q \in \mathbb{R}^{T_q\times d}$ and $K \in \mathbb{R}^{T_k \times d}$ denote the query and key, respectively; $d$ represents the dimension of $Q$ and $K$. We utilize $P_{i,j}$ to represent the associated result of $Q_i \in \mathbb{R}^{d}$ and $K_j \in \mathbb{R}^{d}$. The discrete first derivative of $P$ in the second dimension is obtained as follows:


\begin{equation}
\label{16}
P_{i,j}^{'} = \left\{
\begin{array}{rcl}
|P_{i,j}|&&{j = 0} \\
|P_{i,j} - P_{i,j-1}|&& {j \neq 0}
\end{array} \right.
\end{equation}





For time step $i$ of the query, we choose top-$n$ values in $P_{i}^{'}$, since the boundary of two scenarios always has high gradient value.

\begin{equation}
\label{17}
\mathcal{H}(P, n)_{i,j}=\left\{\begin{array}{ll}
{P_{i,j}} & { P_{i,j}^{'} \geq c_{i}} \\
{-\infty} & { P_{i,j}^{'}<c_{i}}
\end{array}\right.
\end{equation}

\noindent where $c_i$ is the $n$-th largest value of $P_{i}^{'}$. We implement $\rm softmax$ function for the processed $\mathcal{H}(P, n)$.

Furthermore, to keep the time steps with large original logits, we define $P_{i,j}^{*}$ to replace $P_{i,j}^{'}$:

\begin{equation}
\label{18}
P_{i,j}^{*} = \alpha P_{i,j}^{'} + (1-\alpha) P_{i,j}
\end{equation}

$P_{i,j}^{'}$ is a special variant of $P_{i,j}^{*}$ when $\alpha = 1$. 






\subsection{Theoretical Analysis of Boundary-Aware Pooling}
\label{4.2}
Suppose we randomly choose one time step of $Q \in \mathbb{R}^{T_q \times d}$ as query $q \in \mathbb{R}^{d}$, the query $q$ associates $K \in \mathbb{R}^{T_k \times d}$ at all the time steps. The logits of scaled dot-product attention are $[p_1,p_2,...,p_{T_k}] \in \mathbb{R}^{T_k}$.
We calculate the attention weight of each time step as:

\begin{equation}
\label{19}
a_\beta = \frac{\exp(p_\beta)}{\sum_{t_k=1}^{T_k} \exp(p_{t_k})}
\end{equation}

To the best of our knowledge, there are about $3{\text -}5$ scenarios on average in a ten-second video clip at a coarse granularity, like the example in Fig. \ref{fig:redun}. One-second clip usually contains $25$ frames. Therefore, most frames in the same scenario are redundant. Existing methods sample the video to a fixed number of frames or directly reduce the frame rate. Although such methods are effective to some extent, there is still much redundancy in the scenarios that have a large number of time steps. The total attention weights of the scenarios with fewer time steps may be influenced. Specifically, we divide $T_k$ time steps into two groups. The scenario one occupies $T_1$ time steps, the remaining $T_2$ time steps belong to scenario two. Suppose that the features of different time steps in the same scenario are the same, we obtain the total weights of two scenarios as follows:

\begin{equation}
\label{20}
A_{s_\gamma} = \frac{T_\gamma \exp(p_{s_\gamma
	})}{\sum_{o = 1}^{2}T_o \exp(p_{s_o})}, \gamma \in \{1, 2\}
\end{equation}




\noindent where $A_{s_\gamma}$ denotes the total weight of scenario $\gamma$, $p_{s_\gamma}$ denotes the associated logit. Suppose the query is related to scenario two ($p_{s_1} < p_{s_2}$) and $T_1 > T_2$, the ratio of $T_1$ to $T_2$ may influence the total attention weights ($A_{s_1}$ and $A_{s_2}$) of two scenarios.



More concretely, we assume that $T_1$ is $0.75T_k$ and $T_2$ is $0.25T_k$. $A_{s_1}$ and $A_{s_2}$ are calculated as:


\begin{equation}
\label{21}
A_{s_1} = \frac{3\exp(p_{s_1})}{3\exp(p_{s_1}) + \exp(p_{s_2})}
\end{equation}

\begin{equation}
\label{22}
A_{s_2} = \frac{\exp(p_{s_2})}{3\exp(p_{s_1}) + \exp(p_{s_2})}
\end{equation}

\noindent if we apply sparse boundary-aware pooling strategy ($P_{i,j}^{'}$) for the logits and sample one time step in each scenario. Both $A_{s_1}$ and $A_{s_2}$ are transformed and the weight of scenario two obviously increases.

\begin{equation}
\label{23}
A_{s_1}^{'} = \frac{\exp(p_{s_1})}{\exp(p_{s_1}) + \exp(p_{s_2})} < A_{s_1}
\end{equation}

\begin{equation}
\label{24}
A_{s_2}^{'} = \frac{\exp(p_{s_2})}{\exp(p_{s_1}) + \exp(p_{s_2})} > A_{s_2}
\end{equation}

However, when the query is related to scenario one ($p_{s_1} > p_{s_2}$). It is not appropriate to reduce the proportion of scenario one. Therefore, we define $P_{i,j}^{*}$ to replace $P_{i,j}^{'}$ and select not only the boundaries of scenarios, but also the time steps with large original logits. Specifically, the number of selected steps is $n$, we sample two boundaries in the two scenarios and the remaining $n-2$ time steps belong to scenario one. $A_{s_1}^{'}$ is obtained as:

\begin{equation}
\label{25}
A_{s_1}^{'} = \frac{(n-1)\exp(p_{s_1})}{(n-1)\exp(p_{s_1}) + \exp(p_{s_2})}
\end{equation}

\noindent for the increase from $A_{s_1} $ in Eqn. \ref{21} to $A_{s_1}^{'}$ in Eqn. \ref{25}, we just need to ensure that $n-1 > 3$.

When the video clip has more than two scenarios, we also divide them into two groups. One has the scenarios with larger logits, the other has the remaining scenarios. The above analysis of two scenarios is approximately applicable in this situation.


\begin{table*}[!t]
	\renewcommand{\arraystretch}{1.2}
	\setlength\tabcolsep{3.0pt}
	\centering
	\begin{tabular}{c|c|c|c|c|c|c|c|c}
		\hline
		\multirow{2}*{Method} & \multicolumn{4}{c|}{MSVD} & \multicolumn{4}{c}{MSR-VTT} \\
		\cline{2-9}
		& BLEU4
		& ROUGE
		& METEOR
		& CIDEr
		& BLEU4
		& ROUGE
		& METEOR
		& CIDEr\\
		
		\hline\hline
		Vanilla Transformer &51.4 &69.7 &34.6 &86.4 &40.9 &60.4 &28.5 &48.9 \\
		SBAT (w/o CM) &52.4 &71.2 &35.0 &87.0 &42.0 &60.8 &28.5 &50.1 \\
		SBAT (w/o Local) &\bf 53.5 &\bf 72.3 &35.2 &88.9 &41.9 &61.0 &28.4 &50.5 \\
		SBAT (Sample) &51.3 &71.9 &35.2 &88.6 &42.3 &61.0 &28.7 &51.0 \\
		
		SBAT &53.1 &\bf 72.3 &\bf 35.3 &\bf 89.5 &\bf 42.9 &\bf 61.5 &\bf 28.9 &\bf 51.6 \\

		\hline
	\end{tabular}
	\caption{Evaluation results of our proposed methods. Note that we reproduce the results of Vanilla Transformer (TVT [Chen et al., 2018]). Due to different learning rate strategy, our implementation achieves better performances than the original TVT on MSR-VTT.}
	\label{table:compa}
\vspace{-0.3cm}
\end{table*}

\subsection{Local Correlation}
\label{4.3}
Since we employ sparse boundary-aware strategy for the attention logits, the local information between consecutive frames is ignored. We develop a local correlation scheme based on the multihead attention to compensate for the information loss. Formally, the original logits $P$ are obtained following Eqn. \ref{16}. The correlation scheme is
\begin{equation}
\label{26}
\mathcal{H}_{\rm corr}(P, n)_{i,j}=\left\{\begin{array}{ll}
{P_{i,j}} & { |i-j| \leq r} \\
{-\infty} & { |i-j|> r}
\end{array}\right.
\end{equation}

\noindent where $r$ denotes the maximum distance of two frames and the correlation size is $2r$. In practice, the local correlation and boundary-aware correlation are utilized simultaneously. 

\subsection{Cross-Modal Scheme}
\label{4.4}
Existing methods deal with different modalities separately in the encoder and ignore the interaction between different modalities. Here, we propose an aligned cross-modal scheme based on sparse boundary-aware attention. We divide the video into a fixed number of video chunks and then extract image and motion features from these chunks at the same intervals. Therefore, the feature vectors at the same step are extracted from the same video chunk. We directly apply our sparse boundary-aware attention to the aligned features. When the query is image modality, the key is motion modality, vice versa. Taking the former situation as an example, we compute the results of vanilla and boundary-aware cross-modal attentions as follows:  

\begin{equation}
\label{27}
{\rm CM {\text -} Attention}(I,M) = {\rm MultiHead}(I,M,M)
\end{equation}

\vspace{-0.5cm}

\begin{equation}
\label{28}
{\rm Sp{\text-}CM {\text -} Attention}(I,M) = {\rm Sp {\text -} MultiHead}(I,M,M)
\end{equation}

\noindent where CM denotes cross-modal.

\section{Video Captioning with SBAT}

We introduce the encoder-decoder structure combined with our sparse boundary-aware attention for video captioning. As shown in Fig. \ref{fig:overall}(b), we replace all the vanilla multihead attention blocks with boundary-aware attention blocks, except for the self-attention block for target word embeddings. Different from the original structure, an additional cross-modal attention layer is adopted following the self-attention layer in the encoder. In the decoder, we also introduce the boundary-aware attention into the enc-dec attention layer, but we set $\alpha$ to $0$ in Eqn. \ref{18} and do not use local correlation. 

\section{Experimental Methodology} 
\subsection{Datasets and Metrics}
We evaluate SBAT on two benchmark video captioning datasets, MSVD \cite{chen2011collecting} and MSR-VTT \cite{xu2016msr}. Both the datasets are provided by Microsoft Research, and a series of state-of-the-art methods have been proposed based on these datasets in recent years. MSVD contains $1970$ video clips and each video clip is about $10$ to $25$ seconds long and annotated with about $40$ English sentences. MSR-VTT is larger than MSVD with $10000$ YouTube video clips in total and each clip is annotated with $20$ English sentences. We follow the commonly used protocol in the previous work and evaluate methods under four standard metrics including BLEU, ROUGE, METEOR, and CIDEr.

\subsection{Data Preprocessing}
We extract image features and motion features of video data. For image features, we sample video data to $80$ frames and use the pre-trained Inception-ResNet-v2 \cite{szegedy2017inception} model to obtain the activations from the penultimate layer. For motion features, we divide the raw video data into video chunks centered on the sampled frames and use the pre-trained I3D \cite{carreira2017quo} model to obtain the activations from the last convolutional layer. We implement a mean-pooling operation along the temporal dimension to get the motion features. On MSR-VTT, we also employ glove embeddings of the auxiliary video category labels to facilitate feature encoding.


\subsection{Experimental Details}

The hidden size is set to $512$ for all the multihead attention mechanisms. The numbers of heads and attention blocks are $8$ and $4$, respectively. The value of $\alpha$ is set to $0.8$ in the encoder and $0$ in the decoder. In the training phase, we use Adam \cite{kingma2014adam} algorithm to optimize the loss function. The learning rate is initially set to $0.0001$. If the CIDEr on validation set does not improve over $10$ epochs, we change the learning rate to $0.00002$. The batch size is set to $32$. In the testing phase, we use the beam-search method with a beam-width of $5$ to generate words. We use the pre-trained word2vec embeddings to initialize the word vectors. Each word is represented as a $300$-dimension vector.

\vspace{-0.3cm}

\begin{figure}[h]
	\centering
	\includegraphics[scale=0.4]{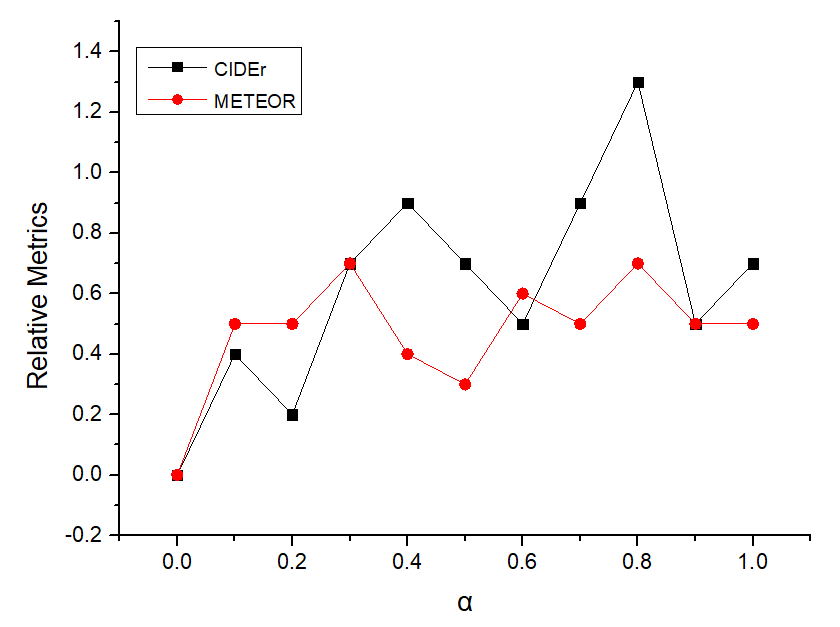}
	\vspace{-0.2cm}
	\caption{Effect of $\alpha$ on MSR-VTT. We show the relative results on METEOR and CIDEr. Specifically, we set $\alpha=0$ as the baseline.} 
	\label{fig:curve} 
	\vspace{-0.3cm}
\end{figure}

\begin{figure*}[h]
	\centering
	\includegraphics[scale=0.7]{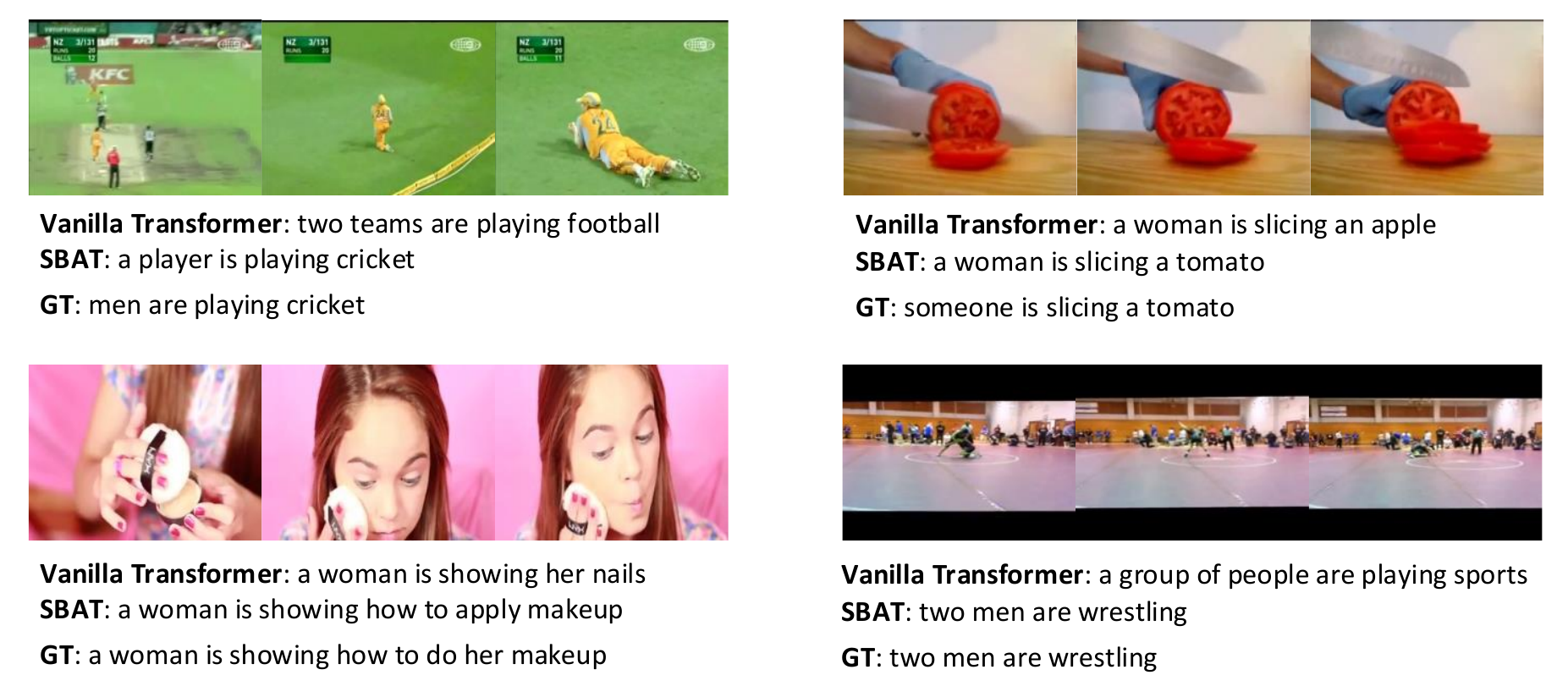}
	\vspace{-0.2cm}
	\caption{Some qualitative results of the video clips on the test sets of MSR-VTT and MSVD. We provide the ground-truth description and the generated descriptions of Vanilla Transformer and SBAT for each video clip.}
	\label{fig:result} 
	\vspace{-0.3cm}
\end{figure*}

\section{Experimental Results}

\subsection{Impact of Sparse Boundary-Aware Attention}

We first evaluate the effectiveness of different variants of SBAT, as shown in Table \ref{table:compa}. Vanilla Transformer and SBAT denote the models in Fig. \ref{fig:overall}(a) and (b). SBAT (w/o CM) denotes the model without aligned cross-modal attention. SBAT (w/o Local) denotes the model without local correlation in the encoder. SBAT (Sample) denotes the model with equidistant sampling for all the time steps, rather than our boundary-aware operation.

In Table \ref{table:compa}, Vanilla Transformer achieves relatively bad results on both datasets. However, when we adopt boundary-aware or equidistant sampling strategies in the multihead attentions, the performances are obviously improved. SBAT with boundary-aware attention, local correlation, and aligned cross-modal interaction achieves promising results under all the metrics. The comparison between SBAT (w/o CM) and SBAT shows that the cross-modal interaction provides useful cues for generating words. The comparison between SBAT (w/o Local) and SBAT shows that the local correlation can make up the loss of local information. Comparing SBAT and SBAT (Sample), although equidistant sampling reduces the feature redundancy to some extent, the ratio between different scenarios is not considered, while SBAT solves this problem effectively. 




\subsection{Comparison of $P^{'}$ and $P$}

To evaluate the impact of $P^{*}$ and find an appropriate ratio between $P^{'}$ and $P$, we adjust the value of $\alpha$ in Eqn. \ref{18} based on SBAT. The experimental results are shown in Fig. \ref{fig:curve}. Note that we only adjust the value of $\alpha$ in the encoder, and the value of $\alpha$ in the decoder is always $0$. We observe that $P^{*}$ with $\alpha = 0.8$ achieves the best performances on both METEOR and CIDEr. In addition, only using original logits $P$ ($\alpha=0$) shows the worst performances, indicating that our proposed boundary-aware strategy $P^{'}$ is a significant boost for the transformer-based video captioning model.

\begin{table}[!h]
	\renewcommand{\arraystretch}{1.2}
	\setlength\tabcolsep{6pt}
	\centering
	\begin{tabular}{p{1cm}|p{1.9cm}<{\centering}|p{0.55cm}<{\centering}|p{0.55cm}<{\centering}|p{0.55cm}<{\centering}|p{0.55cm}<{\centering}}
		\hline
		Dataset&Method & B & R & M & C\\
		\hline
		\multirow{7}{*}{\tabincell{c}{MSR-\\VTT}}
		&TVT &40.1 &61.1  &28.2 &47.7 \\
		
		&MGSA &42.4  &- &27.6  &47.5 \\
		&Dense Cap &41.4  &61.1 &28.3  &48.9 \\
		&MARN &40.4 &60.7 &28.1 &47.1 \\
		&GRU-EVE &38.3  &60.7  &28.4  &48.1  \\
		&POS-CG & 42.0 & 61.1 &  28.1 & 49.0\\
		
		
		\cline{2-6}
		
		&SBAT & \bf 42.9&\bf  61.5&\bf 28.9 &\bf 51.6 \\

		\hline
		\hline
		\multirow{7}{*}{\tabincell{c}{MSVD}}
		&TVT &\bf 53.2 &- &35.2 &86.8 \\
		&SCN &51.1 &- &33.5 &77.7 \\
		
		&MARN &48.6 &71.9 &35.1 &\bf 92.2 \\

		&GRU-EVE &47.9 &71.5 &35.0 &78.1 \\
		&POS-CG &52.5 &71.3 &34.1 &88.7 \\
		\cline{2-6}
		
		&SBAT & 53.1 &\bf 72.3 &\bf 35.3 &89.5 \\
		\hline
	\end{tabular}
	\caption{Evaluation results of video captioning, where B, R, M, C denote BLEU4, ROUGE, METEOR, CIDEr, respectively.}
	\label{table:caption_result}
	\vspace{-0.3cm}
\end{table}

\subsection{Comparison with State-of-the-art}
Table \ref{table:caption_result} shows the results of different methods on MSVD and MSR-VTT. For a fair comparison, we compare SBAT with the methods which also use image features and motion features. The comparison methods include TVT \cite{chen2018tvt}, MGSA \cite{chen2019motion}, Dense Cap \cite{shen2017weakly}, MARN \cite{pei2019memory}, GRU-EVE \cite{aafaq2019spatio}, POS-CG \cite{wang2019controllable}, SCN \cite{gan2017semantic}. In Table \ref{table:caption_result}, SBAT shows better or competitive performances compared with the state-of-the-art methods. On MSR-VTT, SBAT outperforms TVT, MGSA, Dense Cap, MARN, GRU-EVE, POS-CG on all the metrics. In particular, SBAT achieves 51.6\% on CIDEr, making an improvement of 2.6\% over POS-CG. On MSVD, SBAT outperforms SCN, GRU-EVE, POS-CG on all the metrics and has a better overall performance than TVT and MARN.

\begin{figure}[h]
	\centering
	\includegraphics[scale=0.65]{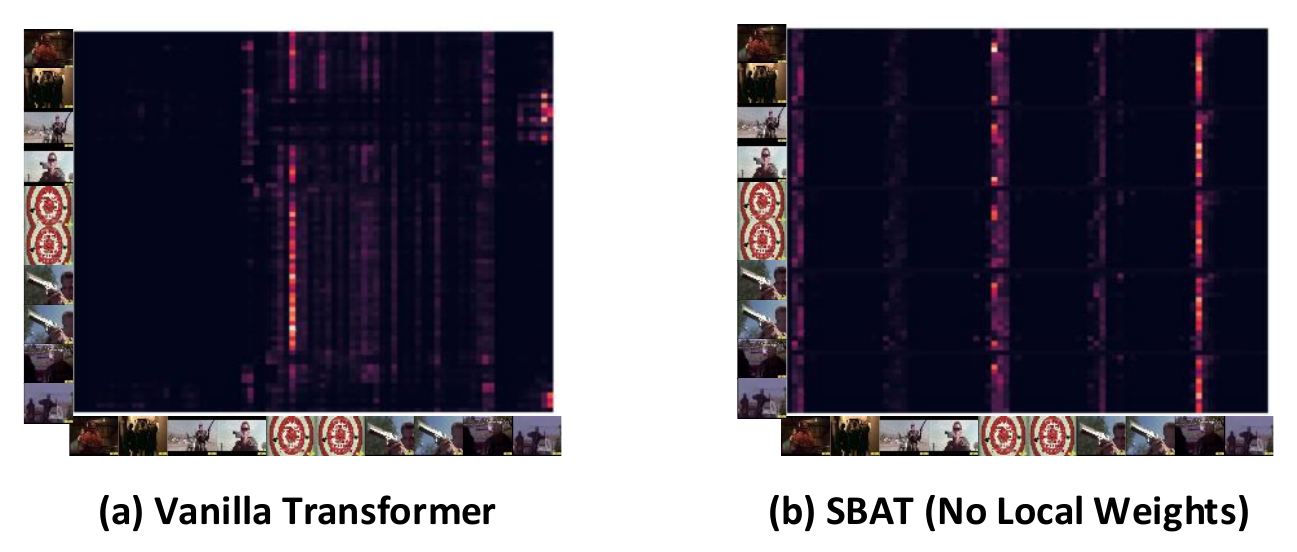}
	\caption{Visualization of attention mechanism. (a) and (b) denote Vanilla Transformer and SBAT, respectively. $x$ and $y$ axes both denote continuous video frames. The generated descriptions of two methods are both ``a man is shooting a gun".}
	\label{fig:result2}
	\vspace{-0.3cm} 
\end{figure}


\subsection{Visualization of Attention Mechanism}

To further illustrate the effectiveness of SBAT, we conduct a case study and visualize the attention distributions of SBAT and Vanilla Transformer. In Fig. \ref{fig:result2}, we take a video clip for example. Note that we only visualize the weights of image modality for convenience, and we do not show the local attention weights. Fig. \ref{fig:result2}(a) shows that the attention weights of Vanilla Transformer are dispersed and Vanilla Transformer has a poor ability to detect the boundary of different scenarios. While Fig. \ref{fig:result2}(b) shows that (1) SBAT has more sparse attention weights than Vanilla Transformer; (2) SBAT accurately detects the scenario boundaries.


\subsection{Qualitative Results}

Fig. \ref{fig:result} shows several qualitative examples. We compare the descriptions generated by Vanilla Transformer, SBAT, and ground truth (GT). With the help of redundancy reduction and a better usage of global and local information, SBAT generates more accurate descriptions that are close to GT.

\section{Conclusion}

In this paper, we have proposed a new method called sparse boundary-aware transformer (SBAT) for video captioning. Specifically, we have proposed sparse boundary-aware strategy for improving the attention logits in vanilla transformer. Combined with local correlation and cross-modal encoding, SBAT can effectively reduce the feature redundancy and capture the global-local video information. The quantitative, qualitative, and ablation experiments on two benchmark datasets have demonstrated the advantage of SBAT.

\section*{Acknowledgments}

This work is supported in part by Science and Technology Innovation 2030 –“New Generation Artificial Intelligence” Major Project No.(2018AAA0100904), National Key R\&D Program of China (No. 2018YFB1403600), NSFC (No. 61672456, 61702448, U19B2043), Artificial Intelligence Research Foundation of Baidu Inc., the funding from HIKVision and Horizon Robotics, and ZJU Converging Media Computing Lab.

\bibliographystyle{named}
\bibliography{ijcai20}

\end{document}